\pdfoutput=1
\documentclass[11pt]{article}

\usepackage[]{acl}

\usepackage{times}
\usepackage{latexsym}

\usepackage[T1]{fontenc}
\usepackage[utf8]{inputenc}

\usepackage{microtype}

\usepackage{inconsolata}

\usepackage{graphicx}
\usepackage{amsmath}
\usepackage{algorithm}
\usepackage{algpseudocode}
\usepackage{multirow} 
\usepackage{amssymb}
\usepackage{textcomp}
\usepackage{colortbl}        % For coloring entire rows/columns if needed
\usepackage{booktabs}        % For professional-looking table borders
\usepackage{xcolor}
\definecolor{lightgreen1}{RGB}{229, 245, 214}
\definecolor{lightgreen2}{RGB}{199, 243, 182}
\definecolor{midgreen}{RGB}{161, 217, 155}
\definecolor{darkgreen1}{RGB}{140, 210, 115}
\definecolor{darkgreen2}{RGB}{70, 160, 65}
\title{Active Learning for Robust and Representative LLM Generation in Safety-Critical Scenarios}

\author{Sabit Hassan\textsuperscript{$\dagger$} Anthony Sicilia\textsuperscript{$\P$} and Malihe Alikhani\textsuperscript{$\P$}\\
  \textsuperscript{$\dagger$}School of Computing and Information, University of Pittsburgh, Pittsburgh, PA, USA  \\
  \textsuperscript{$\P$}Khoury College of Computer Science, Northeastern University, Boston, MA, USA  \\
  \texttt{sabit.hassan@pitt.edu}, \texttt{\{a.sicilia,m.alikhani\}@northeastern.edu} \\
  }

\begin{document}
\maketitle
\begin{abstract}

% It is crucial to ensure that safety measures in user-facing systems are robust against variations. 
% When presented with a random sample of real-world data, LLMs would generate text representing more frequent populations and may not cater to the safety of under-represented groups in data. To address this, we propose a novel active learning-based framework where an LLM's generation is guided by a smaller active learner model. Enhanced with clustering, our method leads to more diverse and fairer set of generations by the LLM without knowing true distribution of data beforehand. We validate our approach by constructing a dataset of \textbf{5.4K} potential safety violations in tandem with LLM generation and training the smaller active learner model simultaneously. Our results demonstrate that the proposed framework not only leads to more representative set of generations but also that the generations lead to more improvement in active learner performance, as well as models outside the active learning loop. We make our code and data publicly available upon acceptance.

% While generating safety-related text with Large Language Models (LLMs), it is crucial to ensure they are representative of different sub-populations in data. 

Ensuring robust safety measures across a wide range of scenarios is crucial for user-facing systems. While Large Language Models (LLMs) can generate valuable data for safety measures, they often exhibit distributional biases, focusing on common scenarios and neglecting rare but critical cases. This can undermine the effectiveness of safety protocols developed using such data. To address this, we propose a novel framework that integrates active learning with clustering to guide LLM generation, enhancing their representativeness and robustness in safety scenarios. We demonstrate the effectiveness of our approach by constructing a dataset of \textbf{5.4K} potential safety violations through an iterative process involving LLM generation and an active learner model's feedback. Our results show that the proposed framework produces a more representative set of safety scenarios without requiring prior knowledge of the underlying data distribution. Additionally, data acquired through our method improves the accuracy and F1 score of both the active learner model as well models outside the scope of active learning process, highlighting its broad applicability.
% Upon acceptance, we will make our code and data publicly available.

% Ensuring robust safety measures across diverse safety scenarios is essential for user-facing systems. While Large Language Models (LLMs) can play a significant role in developing safety-measures by generating data, LLMs may focus on more common scenarios, overlooking rare but critical cases. This distributional bias can be inherited by the safety measures developed using such data, compromising their effectiveness. To address this issue, we propose a novel framework that integrates a smaller active learner model with clustering to customize LLM-generation, making it more robust and representative of safety scenarios. We validate our approach by constructing a dataset of \textbf{5.4K} potential safety violations, created through the interaction between LLM generation and active learner training. Our results demonstrate that this framework yields a more representative set of scenarios without requiring prior knowledge of the true data distribution. Further, our approach also enhances the accuracy and F1 score of the active learner model, as well models outside the scope of active learning process, showcasing the method's versatility and effectiveness. Upon acceptance, we will make our code and data publicly available.
 % We validate the transferability of the approach to a wide array of models, addressing the concerns of using active learning in practice. 
\end{abstract}

\section{Introduction}
% The widespread integration of language models (LLMs) into everyday digital interactions presents challenges in ensuring their safety and robustness. 
% \citet{santy-etal-2023-nlpositionality} highlight biases in NLP datasets and their reflection in model behavior.

% Safety measures in user-facing systems need to be robust against both common and uncommon scenarios. 
LLMs have shown much promise in data generation \cite{radharapu-etal-2023-aart}, which can be leveraged to obtain safety-related data. This data can then be employed to implement safety measures in various models \cite{radharapu-etal-2023-aart,sun-etal-2022-safety}. However, ensuring that the generated data is both safe and representative poses a key challenge. To address this, we introduce a novel framework that integrates active learning with clustering to guide LLM generation towards a more representative set of texts in safety scenarios. 

\begin{figure}
\centering
\includegraphics[width=0.99\linewidth]{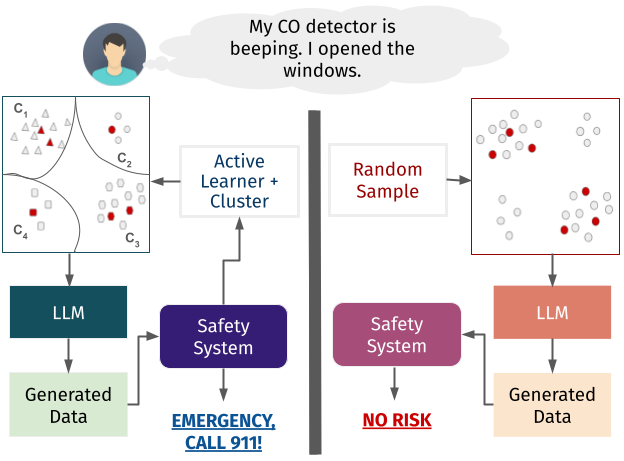}
\caption{Safety systems trained with random LLM generated data may not be resilient against uncommon scenarios. Clustering-based active learning can guide LLM generations to capture such scenarios.}
\label{fig}
\end{figure}
The challenge of making LLM generations both representative and safe arises from inherent distributional biases in real-world data. These biases often cause LLM-generated content to mirror the imbalances, resulting in an over-representation of common scenarios and an under-representation of rare but critical situations. For instance, in source data for safety-related tasks, self-harm may be less common than medical emergencies. Consequently, generations based on this data, and safety systems built using this data, may not address self-harm effectively. Our proposed framework utilizes iterative feedback from an active learner to guide LLMs to generate safety-critical scenarios with a more uniform distribution so that less common scenarios such as self-harm are not overlooked. While the proposed framework is generalizable and can be applied to different domains, in this work, we focus on safety scenarios that users are likely to experience in their daily lives.

% We hypothesize that guiding LLM generation with a clustering-based active learning framework can address these issues resulting from imbalances in data.

% In this framework, the LLM generation and training of the active learner model are interleaved. 
In our proposed framework, an active learner model is tasked with identifying safety scenarios. \textit{Informative instances} for the active learner (i.e., instances the learner is uncertain on) are identified from a \textit{diverse set of regions} of the data represented by different clusters, and are passed to the LLM. The LLM generated output is then used to update the active learner and the process is repeated. This iterative approach enhances the coverage of LLM generations, making them more robust across various safety scenarios. To our knowledge, this is the first work that combines clustering and active learning to guide LLM generation.

% our approach enhances the coverage of LLM-generated outputs, making them more robust across various safety scenarios. 
% In this framework, the LLM functions as a distillation model, with its generations informing the active learner model. 

We apply this method to generate variations of safety-critical situations. Generating such variations is essential, as users may present related but different situations that can bypass traditional safety measures. While previous works have argued for the importance of safety in critical situations \cite{sun-etal-2022-safety,Dinan2021AnticipatingSI}, our approach focuses on generating a diverse and representative array of safety scenarios. By combining various taxonomies of safety situations, we construct a fine-grained dataset using our clustering-based active learning guided LLM generation, resulting in a dataset of \textbf{5.4K} safety violations across six categories. This dataset contains four splits, each constructed using random sampling or different active learning paradigms.

Our results demonstrate that clustering-based active learning leads LLM generation to successfully capture content from less frequent classes \textit{without prior knowledge of the data distribution}. Additionally, safety detection models trained on the data generated with active learner feedback \textit{outperform those trained on other splits and exhibit a more uniform ratio of errors}. We also investigate a key question raised in previous work \cite{lowell-etal-2019-practical}—\textit{whether data acquired by an active learner can be effectively transferred to other models.} Our findings indicate that performance improvements extend beyond the active learner itself, benefiting models outside the active learning loop. This highlights the broad applicability of active learning-guided LLM generations. Our results validate the practical application of active learning by constructing datasets from scratch in tandem with model training, addressing a significant gap in NLP literature \cite{zhang-etal-2022-survey}, where prior work has mainly focused on simulation-based evaluations.

Thus, the contributions of this paper are:
\begin{itemize}
\item A novel framework using clustering and active learning to guide LLMs towards generating safer and more representative outputs in safety scenarios.
\item A publicly available dataset of \textbf{5.4K} safety violations, annotated with a fine-grained taxonomy.
\item Validation of active learning’s performance improvements and transferability  of acquired data in practice, going beyond simulations.
\end{itemize}

We make our dataset publicly available \footnote{Download link for dataset: \url{https://github.com/sabithsn/active-learning-safety}}

\begin{figure*}
    \begin{centering}
  \includegraphics[width=0.80\textwidth]{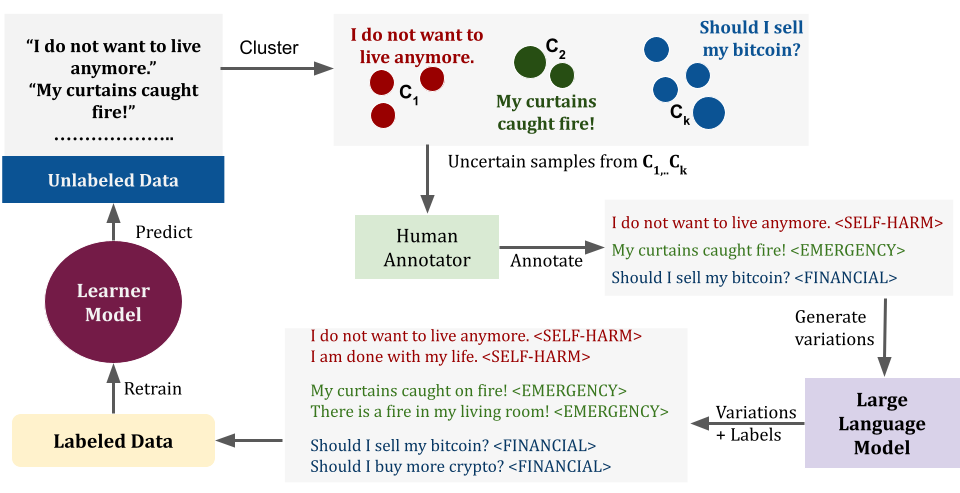}
  \caption{Our proposed framework combines active learning and clustering to guide generations of LLM. Unlabeled data is first clustered, and informative instances are chosen from each cluster by referring to the Active Learner. These instances are then passed to LLM for generation. The active learner is updated at end of each iteration.}
  \label{fig:teaser}
  \end{centering}
\end{figure*}

\section{Related Work}
\paragraph{Active Learning for Language Models}
Active learning is a prominent area in machine learning \cite{Settles2009ActiveLL}, receiving increased attention within NLP \cite{zhang-etal-2022-survey}. Recent applications include active learning with BERT for tasks like intent classification \cite{Zhang2019ensemble}, sentence matching \cite{bai-etal-2020-pre}, and named entity recognition \cite{Liu2022LTPAN}. Innovations include continued pretraining on unlabeled data \cite{margatina-etal-2022-importance} and adaptation to multi-task scenarios \cite{rotman-reichart-2022-multi}. Empirical studies by \citet{ein-dor-etal-2020-active} assess active learning strategies on binary classification. Clustering and advanced active learning strategies are also explored \cite{hassan-alikhani-2023-calm,yuan-etal-2020-cold, margatina-etal-2021-active} for classification tasks.  Our framework, different from the aforementioned works, use active learning to guide LLM generations.

\paragraph{Data Generation with LLMs}
Utilizing LLMs for dataset generation has gained traction \cite{radharapu-etal-2023-aart,chung-etal-2023-increasing,li-etal-2023-synthetic,sicilia2023isabel}, involving tasks from red teaming to emotion classification. The generated data is often used to train other models. For instance, generations from Llama 2 \cite{touvron2023llama} are used to train a classifier which in turn, is used to help training of Llama 3 \cite{llama3modelcard}. Data generation has also been used to train classifier models in Reinforcement Learning with Human Feedback systems \cite{bai2023qwen}. Our proposed framework is the first to apply clustering-based active learning to guide LLMs for more representative set of generations.

\paragraph{AI Safety}
AI safety discussions are prevalent, with frameworks emerging to address risks associated with language models \cite{Dinan2021AnticipatingSI,sun-etal-2022-safety,Weidinger2022TaxonomyOR}. Bias is a significant concern, with efforts to mitigate specific biases, such as gender bias \cite{Lu2020GenderBI,ahn-oh-2021-mitigating,sap-etal-2019-risk}. Other works  often rely on availability of large amount of data for rebalancing or re-annotation \cite{sap-etal-2019-risk,han-etal-2022-balancing}. Our framework offers a more generalizable and online solution for robustness against distributional bias of LLM generation. Our work also contributes a publicly available dataset focusing on fine-grained safety scenarios and safety variations for which there is still a lack of publicly available resources \cite{Dinan2021AnticipatingSI}.

\section{Framework}

% Infusing clustering with active learning can address this problem while mitigating bias overall as shown in \cite{hassan-alikhani-2023-calm}. 
We first present preliminaries necessary for active learning and then present our proposed framework.

\subsection{Preliminaries} 
% Here we describe preliminaries necessary for active learning.
\paragraph{Labeling Scenario} We assume there is a large pool of unlabeled dataset \textit{U} but, expanding on standard active learning, only a subset of labeled data \textit{L} can be used for generation. \textit{L} is iteratively constructed by querying generated output for the \textit{most-informative} instance. While other active learning scenarios exist \cite{Settles2009ActiveLL}, we follow the setting of \textit{pool-based} active learning because of its relevance to many recent NLP tasks for which a large amount of unlabeled data is scraped from the web and then a subset of it is annotated.

\paragraph{Query-Strategy} Different query-strategies have been proposed for identifying relevant instances in active learning, with uncertainty based sampling being the most popular one. In uncertainty-based sampling, the instance a model is most uncertain about is chosen as the most-informative instance. The most commonly used measure of uncertainty is entropy \cite{Settles2009ActiveLL}:

\begin{equation}
\label{Entropy}
    x_{E}^* = \underset{x}{argmax} \; - \underset{i}{\sum} P_\theta(y_i|x)logP_\theta(y_i|x)
\end{equation}
In Eq. \ref{Entropy}, i ranges over all possible labels. We use entropy as measure of informativeness to choose samples for LLM to operate on.
% In standard active learning, a learner model is bootstrapped with some data, then the model is trained iteratively by querying labels for most informative instances and retraining the model after each iteration. 

\subsection{Clustering-based Active Learning guided LLM Generation} 
Active learning typically identifies highly informative instances by measuring uncertainty, such as entropy \cite{Settles2009ActiveLL}. It can induce biased behavior if the model misjudges its confidence \cite{hassan2018interactive}. Clustering, which naturally garners diverse samples \cite{yuan-etal-2020-cold}, combined with active learning, can counteract this by simultaneously gathering diverse and informative data. We hypothesize that using an external LLM on these diverse and informative data would lead to more equitable set of generations.

In our clustering-based setting, the unlabeled data is first vectorized and then the vector space is split into $m$ clusters $\{C_1,C_2,...C_m\}$ where $m$ is a predefined number. Uncertainty measure (e.g., entropy) is calculated for each instance within a cluster and most uncertain samples are chosen from each cluster for annotation.

% \subsubsection{Clustering-based AL with Knowledge Distillation} Having covered the preliminaries, we are now ready to integrate knowledge distillation with clustering-based active learning. Algorithm \ref{alg:cap} summarizes our proposed approach.

% \begin{algorithm}
% \caption{Knowledge distilled Clustering-Based Active Learning}\label{alg:cap}
% \begin{algorithmic}
% % \Require $n \geq 0$
% % \Ensure $y = x^n$
% \State $U,L \gets$ unlabeled data, labeled data
% % \State $L \gets$ labeled data
% \State $S \gets$ LLM for distillation
% \State $G \gets$ bootstrapped model
% \State $B \gets$ labeling budget
% \State $N \gets$ annotation batch size
% \State $m \gets$ initial number of clusters
% \State $V \gets$ vectorize U
% \State Cluster $V$ into \{$C_1$, $C_2$, ... $C_m$\}
% \While{$B \geq 0$}
%     % \State $N=N-k$

%     \For{\texttt{i=0,1,...m}}
%         \For{\texttt{j=0,1,...$|C_i|$}}
%             \State $E_{ij} \gets$ Entropy(G,$X_{ij}$)
%         \EndFor
    
%         \State $x_{i}^* \gets \underset{j}{argmax}(E_{ij}$)
%         \State $y_{i}^* \gets$ generation template T for $x_{i}^*$
%         % \State Add $(x_{i}^*, y_{i}^*)$ to $L$
%         \State $\{(x_{ik}^*, y_{i}^*)\} \gets$ Distill S with $T(x_{i})$
%         \State Add $\{(x_{ik}^*, y_{i}^*)\}$ to $L$ 
%     \EndFor
%     \State $G \gets$ retrain on $L$
%     \State $B=B-N$
% \EndWhile
% \end{algorithmic}
% \end{algorithm}
In standard active learning a human annotator would label this set of samples. In our framework, we assume we have access to an LLM, $S$, and we want to leverage generation of $S$ with respect to informative instances of learner model $G$.
To do so, we introduce concept of a \textit{template}. A \textit{template} $T$ is a prompting structure to guide the generation of the LLM $S$:
\begin{quote}
    \textit{$T(x, O(x))$ : on input $x$, prompt $S$ to generate $\{f(x_1), f(x_2),...f(x_k)\}$ such that $R(f(x_i),O(x))$ holds.} 
\end{quote}
Here, we define $f(x_i)$ to be a variation of input $x$, k as the number of variations we want, and $R(f(x_i),O(x))$ is a relation that evaluates to $True$ if the label for $f(x_i)$ matches the human label $O(x)$ for input x. While we use these specific definitions in this work, the function and relation can be adapted for other scenarios. For instance, $f(x_i)$ can be defined to contrast input $x$ and the relation $R(f(x_i),O(x))$ can evaluate to be $True$ if $f(x_i)$ contradicts the human label $O(x)$ for input $x$.

% $f(x)$ is the target task, such as "identify risk category". And $L$ is instruction specific to the task. $L$ can be in different forms depending on the task. For instance, $L(x)$ can be "generate variations of input x".

% Entropy is used to identify informative instances for classification tasks. 

% $L(x)$ can be more complex, "apply $f(x)$ while respecting definition of social politeness: [politeness definition]".  

\begin{algorithm}
\caption{Active Learner Guided Generation}\label{alg:nlg-al}
\begin{algorithmic}
% \Require $n \geq 0$
% \Ensure $y = x^n$
\State $U,L \gets$ unlabeled data, labeled data
% \State $L \gets$ labeled data
\State $S \gets$ LLM for distillation
\State $G \gets$ bootstrapped model
\State $B \gets$ labeling budget
\State $N \gets$ annotation batch size
\State $m \gets$ number of clusters
\State $V \gets$ vectorize U
\State $O \gets$ human annotator
\State Cluster $V$ into \{$C_1$, $C_2$, ... $C_m$\} 
\While{$B \geq 0$}
    % \State $N=N-k$

    \For{\texttt{i=0,1,...m}}
        \For{\texttt{j=0,1,...$|C_i|$}}
            \State $E_{ij} \gets$ Entropy($x_{ij})$  
        \EndFor
    
        \State $x_{i}^* \gets \underset{j}{argmax}(E_{ij}$) 
        \State $y_{i}^* \gets$ Annotate $O((x_{i}^*)$ 
        \State $T_{i}^* \gets$ generation template T for $x_{i}^*$
        % \State Add $(x_{i}^*, y_{i}^*)$ to $L$
        
        \State $\{(x_{ik}^*, y_{ik}^*)\} \gets$ Distill S, $T_i(x_{i}^*, O(x_i^*))$
        % \State Validate $O(\{(x_{ik}^*, y_{i}^*)\}$ 
        \State Add $(x_i^*,y_i^*)$ and $\{(x_{ik}^*, y_{i}^*)\}$ to $L$ 
    \EndFor
    \State $G \gets$ retrain on $L$
    \State $B=B-N$
\EndWhile
\end{algorithmic}
\end{algorithm}

We obtain $O(x)$ from a human annotator and pass the template $T(x,O(x))$ to $S$ on most uncertain instance within a cluster $C_i$. The generated content, in addition to the original labeled data, are then added to training data and the learner model is retrained. This process continues iteratively until resources run out. We present our approach formally in algorithm \ref{alg:nlg-al}.

 % The algorithm starts with an initial boot-strapped model that we wish to train and a large model such as GPT-3.5 that we distill knowledge from using the template described earlier. The boot-strapped model is trained iteratively on informative instances from each cluster until labeling resources run out.

% The algorithm starts with an initial boot-strapped model that we wish to train and a large model such as GPT-3.5 that we distill knowledge from using a template described earlier. The boot-strapped model is trained iteratively on informative instances from each cluster until labeling resources run out. Entropy is used to identify informative instances for classification tasks. 

\section{Dataset}
\subsection{Taxonomy} We combine existing categorization \cite{dinan2021anticipating,sun-etal-2022-safety,Weidinger2022TaxonomyOR} of safety into a unified taxonomy. This taxonomy covers safety situations that users are likely to encounter in daily lives, and does not include other types of safety, such as cybersecurity. The taxonomy covers six classes:

\paragraph{Self-harm:} Due to the openness of users discussing mental health with chatbots \cite{dinan2021anticipating}, detecting self-harm intentions and preventing harmful response is crucial.
\paragraph{Medical Scenario:} Despite advancements in medical NLP \cite{michalopoulos-etal-2021-umlsbert}, ethical concerns persist \cite{Palanica2018PhysiciansPO}. General LLMs should avoid providing medical advice.
\paragraph{Legal Scenario:} Offering legal advice demands specialized, context-dependent legal knowledge \cite{Susskind2013TomorrowsL}. LLMs, lacking continuous adaptation, should not provide legal advice.
\paragraph{Financial Scenario:} Financial advice requires deep understanding and accountability \cite{graham2003intelligent}, and should be avoided by LLMs unless specialized to do so.
\paragraph{Emergency Scenario:} Non-medical emergencies such as fires or crimes require specific responses \cite{dinan2021anticipating, Chan2019QuestionansweringDS}, unsuitable for general LLM response.

\paragraph{Not Harmful:} No plausible safety concerns.

% In addition, we include a \textbf{"Not-Harmful"} category, resulting in six total categories in our dataset.

\begin{table*}[!h]
    \small
    \centering

    \begin{tabular}{l|p{1.7cm}|p{1.8cm}|p{1.7cm}|p{1.5cm}|p{1.8cm}}
         \textbf{Ref} & \textbf{Self-harm} & \textbf{Medical} & \textbf{Legal} & \textbf{Financial} & \textbf{Emergency} \\
         \hline
         \citet{dinan2021anticipating} & \checkmark & \checkmark & X & X & \checkmark\\
         % Ahichten et al. \cite{Agichtein2023Taskbot}
         % & \checkmark & \checkmark & \checkmark & \checkmark & X\\
         \citet{Sun2021OnTS} & \checkmark & \checkmark & \checkmark & X & X
         \\
        \citet{Weidinger2022TaxonomyOR} & X & \checkmark & \checkmark & X & X\\
         % \hline
         \hline
         Ours & \checkmark & \checkmark & \checkmark & \checkmark & \checkmark\\
         
    \end{tabular}
        \caption{ Our taxonomy of safety-critical situations unifies different categories proposed by prior work. While previous works list these categories as part of risks (e.g., offending user \cite{sun-etal-2022-safety}) with language models, we focus on unifying a fine-grained taxonomy of safety-critical situations.}
    \label{tab:advice-taxonomy}
\end{table*}

\begin{table*}[h]
    \centering
    \small
    \begin{tabular}{p{2cm}|p{5.7cm}|p{6.8cm}}
    \hline
    \textbf{Category} & \textbf{Reddit Text} & \textbf{LLM Generated Variation}\\
    \hline
    \hline
    \multirow{2}*{Self-Harm}	& \multirow{2}*{Feeling depressed again} 	&Struggling with self-destructive thoughts\\
    \cline{3-3}
    & & Emotional pain pushing me to hurt myself\\
    \hline
{Medical}	& \multirow{2}*{What is this rash?}	& What could be causing this skin irritation?	\\
\cline{3-3}
	Advice& & Is this skin reddening normal?	\\
 
 \hline
{Legal} & \multirow{2}*{Motion to balance out child support}	&Can I contest a child support order?\\
\cline{3-3}
		Advice & & Procedure to adjust child support payments.\\
  \hline
{Financial} & \multirow{3}*{I should just pay off my loans, right?}	 
& Should I immediately settle my credit dues?\\
\cline{3-3}
	Advice& & Does discharging my loans early make financial sense?\\	
\hline
Emergency-	& {Where to acquire nerve agent antidote}	& How can I get EpiPen for severe allergies?\\	\cline{3-3}
Situation		& pens? & Where can I find radiation poisoning antidotes?\\	
\hline
\multirow{2}*{Not-Harmful}	& \multirow{2}*{Status check}	& Can I get an update on the status?\\	\cline{3-3}
		& & Update on progress?\\	

    \hline
    \end{tabular}
            \caption{Examples of safety violations from our dataset. Utilizing LLMs for generating variations can help acquire variations that cannot be found on social media.}
\end{table*}

\begin{table*}[!h]
\label{data-dist}
    \centering
    \small
    \begin{tabular}{l||c|c|c|c||c|c|c||c}
    \hline
    & \textbf{Random} & \textbf{TopN} & \textbf{Coreset} & \textbf{Cluster} & \textbf{Bootstrap} & \textbf{Dev} & \textbf{Test} & \textbf{Total}\\
    \hline
    \hline
\textbf{Self-Harm}	& 96 & 116 & 66	& 115 & 22	& 26 & 438 & 879\\
\hline
\textbf{Medical-Advice}	& 180	& 88 & 115 & 121	& 24 & 26 & 474	& 1028\\
\hline
\textbf{Legal-Advice} & 84 & 90 & 137 & 87 & 36	& 32 & 500 & 966\\
\hline
\textbf{Financial-Advice} & 84 & 112 & 90 & 94 &	25 & 29 & 497 & 931\\
\hline
\textbf{Emergency-Situation}	& 12 & 24 & 0 & 30 &	5	& 6  & 82 & 159\\
\hline
\textbf{Not-Harmful}	& 144 & 170	& 192	& 153	& 38 & 31	& 709 & 1437\\
\hline
\hline
\textbf{Standard Deviation}	& \textbf{57.6} &	\textbf{47.6} &	\textbf{65.3} &	\textbf{41.4} &	- & - & - & -\\
\hline
\textbf{Total}	& \textbf{600}	& \textbf{600}	& \textbf{600}	& \textbf{600} & \textbf{150}	& \textbf{150} & \textbf{2700} & \textbf{5400}\\
    \hline
    \end{tabular}
            \caption{Distribution of different categories across splits. Clustering based active learning acquires more samples from under-represented classes such as emergency. Lower standard deviation of counts also indicate reduced bias.}
            \label{tab:data-dist}
\end{table*}
\subsection{Dataset Construction}
As social media can be a diverse source of data \cite{ye-etal-2023-multilingual}, we compile an initial unlabeled pool of data from Reddit. We select posts relevant to five categories of safety-critical situations from 15 subreddits, such as r/depression for self-harm and r/LegalAdvice for legal scenarios, collecting up to 1000 posts per subreddit, totaling \textasciitilde14,000 posts. 

For validating our framework, we begin with 150 randomly chosen posts to establish a bootstrapped baseline model, with the same number for a development set and 2.7K for a more comprehensive test set. This data is manually annotated by two graduate students to ensure relevance to the categories, with an inter-annotator agreement of $\kappa$ 81.89, reflecting high consensus. This setup leaves \textasciitilde 11K posts in the unlabeled pool. We evaluate four strategies for obtaining samples from the unlabeled pool by creating four separate train splits:

\paragraph{Random:} Samples are chosen randomly.
\paragraph{TopN-AL:} Adding the N most informative posts to the training set in each iteration.
\paragraph{Coreset-AL:} Selecting a subset that is representative of the dataset \cite{sener2018active}.
\paragraph{Cluster-AL:} Selecting $N/m$ most-informative posts from each cluster in each iteration.

% Random sampling serves as baseline. Cluster-AL is the paradigm of our framework. TopN-AL demonstrates flexibility of our approach to other AL paradigms. To evaluate these different paradigms consistently, we keep the test set fixed. 
% Algorithm \ref{alg:cap} with data augmentation for templating. We compare three active learning approaches with random sampling:

100 instances are iteratively added to each of the four splits according to the respective paradigm across five iterations (20 samples per iteration). A learner model is used to obtain the most-informative instances. These instances are labeled by a human annotator at each iteration. During each iteration, we generate five variations for each of these newly added instances while respecting the human labels by using our concept of template with the LLM GPT-3.5-turbo\footnote{https://platform.openai.com/docs/models/gpt-3-5}. This yields a total of 600 training instances for each split. Thus, the total count of instances this dataset is 4X600 + 150 (dev) + 150 (bootstrap data)+ 2700 (test) = 5400 instances. 

Critically, we observe in Table \ref{tab:data-dist} that clustering-based active learning acquires more data for low-frequency classes in source data such as "emergency" and also has substantially \textbf{lower standard deviation (41.4 as opposed to 57.6 by random sampling)} of counts per class. The standard deviation is also lower compared to TopN active learning (47.4) and Coreset (65.3) as well. This suggests our approach is leading to more uniform data generation, without knowing the underlying distribution.

\section{Experiments}
We evaluate the quality of LLM generations by evaluating models trained on the generated data. 
\subsection{Models}
% We experiment with traditional SVMs for their lightweight properties and recent transformer models.

% \paragraph{SVMs:} 
We choose a set of small pretrained transformer-based language models fine-tuned with the different data splits in Table \ref{tab:data-dist} to assess the relative efficacy of the different approaches. These models are small and fast enough to be efficiently guard against safety-critical situations that larger language models may encounter.

We use a bert-base-cased \cite{Devlin2019BERTPO} as our learner model. We evaluate transferability of data acquired to four other transformer models, namely: i) bert-base-uncased \cite{Devlin2019BERTPO}, ii) roberta-base \cite{roberta-base}, iii) distilbert-base-cased \cite{Sanh2019DistilBERTAD}, and iv) distilbert-base-uncased \cite{Sanh2019DistilBERTAD}. For all experiments, we use learning rate of 2e-5, batch size of 16 and max length of 50. 

% These models serve the purpose of validating if instances acquired by the learner models can aid other models.

  % We experiment with four transformer models, namely: i) bert-base-c  The bert-base model is used as the active learner. The other models are used to determine if data acquired by the active learner is transferable to other models.

\subsection{Experiment Scenarios}

\paragraph{Baseline classification}
We train our set of models just on the dataset for bootstrapping the models. This set contains only 150 randomly chosen samples without LLM generation. As such, low performance is expected.
\paragraph{Active learning without LLM generation}
We use 100 human labels obtained through random sampling or active learning paradigms in addition to the 150 bootstrapping data.  

\paragraph{Active gearning with LLM generation} We use 500 LLM generated variations along with the human labels and bootstrapping data. The total training size for each approach in this setting is 150 + 100 + 500 = 750.

\subsection{Results}
We use macro-averaged F1 score as primary metric for comparison as the data is imbalanced and this score would provide a better representation of how the models perform on imbalanced data. We also report accuracy, and macro-averaged precision and recall in Tables \ref{tab:stage1-classifier}, \ref{tab:stage2-classifier}, and \ref{tab:stage3-classifier}.

\begin{table}[h]
    \small
    \centering
    \begin{tabular}{l|c|c|c|c}
    \hline
    \textbf{Model} & \textbf{Acc.} & \textbf{Prec.} & \textbf{Rec.} & \textbf{F1}\\
    \hline
    
    bert-base-cased	& 51.8	& 56.1	& 43.1	& 40.7\\
bert-base-uncased	& 46.2	& 46.5	& 37.8	& 36.7\\
roberta-base	& \textbf{72.6}	&\textbf{62.9}	& \textbf{62.3}	&\textbf{61.6}\\
distilbert-base-cased	&35.8	&59.3	&27.7	&19.0\\
distilbert-base-uncased	&68.4	&66.6	&56.3	&57.1\\
% albert-base-v2	&26.3	&4.4	&16.7	&6.9\\
% SVM-char-[2-5]	&43.7	&47.9	&37.0	&36.9\\
% SVM-word-[1-2]	&41.3	&38.0	&34.2	&33.9\\

    \hline
    \end{tabular}
       \caption{Results for identifying safety-violation scenarios prior to active learning and LLM generation. Roberta-base achieves highest results. Other models perform poorly due to very small amount of data.}

    \label{tab:stage1-classifier}
\end{table}
\begin{table*}[ht]
\centering
\begin{tabular}{l|l|c|c|c|c}
\hline
\textbf{Approach} & \textbf{Model} & \textbf{Accuracy} & \textbf{Precision} & \textbf{Recall} & \textbf{F1} \\
\hline
 & bert-base-cased & \cellcolor{lightgreen1} 51.9 & \cellcolor{lightgreen1} 49.5 & \cellcolor{lightgreen1} 46.1 & \cellcolor{lightgreen1} 43.2 \\
% \hline
 & bert-base-uncased & \cellcolor{midgreen} 62.4 & \cellcolor{lightgreen2} 55.9 & \cellcolor{lightgreen2} 53.8 & \cellcolor{midgreen} 52.7 \\
% \hline
Random & roberta-base & \cellcolor{darkgreen2} 75.6 & \cellcolor{darkgreen1} 63.3 & \cellcolor{darkgreen1} 66.0 & \cellcolor{darkgreen2} 64.2 \\
% \hline
 & disbert-base-cased & \cellcolor{darkgreen1} 70.3 & \cellcolor{midgreen} 60.5 & \cellcolor{midgreen} 60.1 & \cellcolor{darkgreen2} 59.3 \\
% \hline
 & disbert-base-uncased & \cellcolor{lightgreen2} 56.9 & \cellcolor{darkgreen1} 61.9 & \cellcolor{lightgreen1} 46.0 & \cellcolor{lightgreen1} 40.8 \\
\hline
\hline
 & bert-base-cased & \cellcolor{lightgreen1} 48.9 & \cellcolor{lightgreen1} 48.7 & \cellcolor{lightgreen1} 45.8 & \cellcolor{lightgreen1} 38.9 \\

 & bert-base-uncased & \cellcolor{darkgreen1} 66.0 & \cellcolor{lightgreen2} 55.1 & \cellcolor{midgreen} 59.0 & \cellcolor{darkgreen1} 55.8 \\

TopN-AL & roberta-base & \cellcolor{darkgreen2} 75.4 & \cellcolor{darkgreen2} 68.8 & \cellcolor{darkgreen1} 67.6 & \cellcolor{darkgreen2} 64.2 \\
 & disbert-base-cased & \cellcolor{darkgreen1} 65.4 & \cellcolor{darkgreen1} 61.9 & \cellcolor{midgreen} 59.4 & \cellcolor{darkgreen1} 57.2 \\
 & disbert-base-uncased & \cellcolor{midgreen} 63.3 & \cellcolor{lightgreen2} 56.1 & \cellcolor{midgreen} 58.7 & \cellcolor{midgreen} 52.0 \\
\hline
\hline
 & bert-base-cased & \cellcolor{lightgreen2} 54.8 & \cellcolor{darkgreen1} 62.3 & \cellcolor{lightgreen1} 44.0 & \cellcolor{lightgreen1} 38.7 \\
 & bert-base-uncased & \cellcolor{lightgreen2} 57.6 & \cellcolor{lightgreen2} 51.6 & \cellcolor{lightgreen2} 49.8 & \cellcolor{lightgreen2} 46.9 \\
Coreset-AL & roberta-base & \cellcolor{darkgreen2} 75.3 & \cellcolor{darkgreen1} 64.8 & \cellcolor{darkgreen1} 64.4 & \cellcolor{darkgreen2} 63.7 \\
 & disbert-base-cased & \cellcolor{darkgreen2} 72.4 & \cellcolor{darkgreen1} 64.1 & \cellcolor{darkgreen1} 61.6 & \cellcolor{darkgreen2} 61.8 \\
 & disbert-base-uncased & \cellcolor{lightgreen2} 58.1 & \cellcolor{darkgreen1} 61.3 & \cellcolor{lightgreen1} 47.2 & \cellcolor{lightgreen1} 41.3 \\
\hline
\hline
 & bert-base-cased & \cellcolor{lightgreen2} 58.6 & \cellcolor{lightgreen2} 51.8 & \cellcolor{lightgreen2} 51.4 & \cellcolor{midgreen} 49.7 \\

 & bert-base-uncased & \cellcolor{darkgreen1} 64.1 & \cellcolor{midgreen} 57.5 & \cellcolor{midgreen} 58.7 & \cellcolor{darkgreen1} 55.8 \\
Cluster-AL & roberta-base & \cellcolor{darkgreen2} 70.6 & \cellcolor{darkgreen2} 67.4 & \cellcolor{darkgreen2} 71.1 & \cellcolor{darkgreen2} 64.3 \\
 & disbert-base-cased & \cellcolor{darkgreen1} 69.6 & \cellcolor{darkgreen1} 63.7 & \cellcolor{darkgreen1} 61.9 & \cellcolor{darkgreen2} 59.4 \\
 & disbert-base-uncased & \cellcolor{midgreen} 61.1 & \cellcolor{lightgreen2} 53.7 & \cellcolor{midgreen} 56.2 & \cellcolor{midgreen} 50.0 \\
\hline
\end{tabular}
\caption{Results for active learning without LLM generation. Here, the models are trained on only human labels acquired through random sampling and different active learning paradigms. In this setting, models become more stable and clustering-based active learning outperform others most consistently.}
        \label{tab:stage2-classifier}

\end{table*}

\begin{table*}[!h]
\centering
\begin{tabular}{l|l|c|c|c|c}
\hline
\textbf{Approach} & \textbf{Model} & \textbf{Accuracy} & \textbf{Precision} & \textbf{Recall} & \textbf{F1} \\
\hline
 & bert-base-cased & \cellcolor{lightgreen2} 74.3 & \cellcolor{darkgreen1} 79.7 & \cellcolor{midgreen} 64.9 & \cellcolor{lightgreen2} 63.7 \\
Random & bert-base-uncased & \cellcolor{darkgreen1} 77.3 & \cellcolor{lightgreen1} 65.5 & \cellcolor{darkgreen1} 67.2 & \cellcolor{darkgreen1} 66.0 \\
+ & roberta-base & \cellcolor{darkgreen1} 78.9 & \cellcolor{lightgreen2} 66.7 & \cellcolor{darkgreen1} 68.0 & \cellcolor{darkgreen1} 67.2 \\
LLM & distilbert-base-cased & \cellcolor{lightgreen2} 74.6 & \cellcolor{lightgreen1} 63.1 & \cellcolor{lightgreen1} 56.5 & \cellcolor{lightgreen1} 57.5 \\
 & distilbert-base-uncased & \cellcolor{darkgreen1} 76.8 & \cellcolor{lightgreen1} 64.8 & \cellcolor{darkgreen1} 66.5 & \cellcolor{midgreen} 65.4 \\
 \hline
 \hline
 & bert-base-cased & \cellcolor{lightgreen2} 74.0 & \cellcolor{lightgreen1} 62.6 & \cellcolor{midgreen} 64.1 & \cellcolor{lightgreen2} 63.2 \\
TopN & bert-base-uncased & \cellcolor{darkgreen1} 76.8 & \cellcolor{lightgreen1} 64.3 & \cellcolor{darkgreen1} 66.7 & \cellcolor{midgreen} 65.4 \\
+ & roberta-base & \cellcolor{darkgreen2} 79.2 & \cellcolor{midgreen} 71.8 & \cellcolor{darkgreen2} 69.3 & \cellcolor{darkgreen1} 68.2 \\
LLM & disbert-base-cased & \cellcolor{lightgreen1} 73.8 & \cellcolor{darkgreen2} 80.0 & \cellcolor{midgreen} 63.6 & \cellcolor{lightgreen2} 63.9 \\
 & disbert-base-uncased & \cellcolor{darkgreen1} 78.1 & \cellcolor{lightgreen1} 65.3 & \cellcolor{darkgreen1} 67.5 & \cellcolor{darkgreen1} 66.3 \\
\hline
\hline
 & bert-base-cased & \cellcolor{darkgreen1} 77.6 & \cellcolor{lightgreen1} 65.7 & \cellcolor{darkgreen1} 66.8 & \cellcolor{darkgreen1} 66.1 \\
Coreset & bert-base-uncased & \cellcolor{darkgreen1} 78.1 & \cellcolor{lightgreen2} 66.6 & \cellcolor{darkgreen1} 67.0 & \cellcolor{darkgreen1} 66.5 \\
+ & roberta-base & \cellcolor{darkgreen1} 77.7 & \cellcolor{lightgreen2} 66.5 & \cellcolor{darkgreen1} 66.7 & \cellcolor{darkgreen1} 66.3 \\
LLM & disbert-base-cased & \cellcolor{lightgreen1} 73.8 & \cellcolor{lightgreen1} 64.2 & \cellcolor{midgreen} 63.3 & \cellcolor{lightgreen2} 63.4 \\
 & disbert-base-uncased & \cellcolor{darkgreen1} 77.3 & \cellcolor{lightgreen1} 66.3 & \cellcolor{darkgreen1} 66.1 & \cellcolor{midgreen} 65.8 \\
\hline
\hline
 & bert-base-cased & \cellcolor{darkgreen1} 77.2 & \cellcolor{darkgreen2} 81.2 & \cellcolor{darkgreen1} 67.3 & \cellcolor{darkgreen1} 66.3 \\
Cluster-AL & bert-base-uncased & \cellcolor{darkgreen1} 77.0 & \cellcolor{lightgreen1} 64.7 & \cellcolor{darkgreen1} 67.2 & \cellcolor{midgreen} 65.6 \\
+ & roberta-base & \cellcolor{darkgreen2} 79.5 & \cellcolor{darkgreen1} 76.5 & \cellcolor{darkgreen2} 71.8 & \cellcolor{darkgreen2} 71.6 \\
LLM & disbert-base-cased & \cellcolor{lightgreen1} 72.4 & \cellcolor{lightgreen2} 69.4 & \cellcolor{midgreen} 65.5 & \cellcolor{midgreen} 65.1 \\
 & disbert-base-uncased & \cellcolor{darkgreen1} 77.9 & \cellcolor{midgreen} 73.1 & \cellcolor{darkgreen2} 69.4 & \cellcolor{darkgreen2} 70.0 \\
 \hline
\end{tabular}
\caption{Results of active learning with LLM generation. Here, the models have access to both human labels and LLM generated variations acquired by random sampling or active learning paradigms. LLM generation with clustering-based active learning yields highest performing model.}
\label{tab:stage3-classifier}
\end{table*}

\begin{table*}[!h]
    \small

    \centering
    \begin{tabular}{l|l|c|c}
    \hline
    % \multicolumn{5}{c}{\textbf{Random}}\\
    \hline
    \textbf{Approach} & \textbf{Input Text} & \textbf{True Label} & \textbf{Predicted Label}\\
    \hline
    \hline

\multirow{2}{1.5cm}{Random + LLM} &
    Sites for current flu, Covid etc? & Not-Harmful & Medical-Advice\\
    &Well, I did the thing. & Not-Harmful & Self-Harm\\
    \hline
    \hline
\multirow{2}{1.3cm}{TopN-AL + LLM}
& Can I get any backlash over \$45? & Legal-Advice & Financial-Advice\\
& Should I open a Certificate of Deposit? & Financial-Advice & Legal-Advice\\
\hline
\hline
\multirow{2}{1.5cm}{Coreset-AL + LLM} 
& I’ve lived everything I want to live & Self-Harm & Not-Harmful\\
& NY state employer health insurance & Legal-Advice & Medical-Advice\\
\hline
\hline
\multirow{2}{1.5cm}{Cluster-AL + LLM} 
& Can I Learn to Like Exercise? & Not-Harmful & Self-Harm\\
& \$25k unexpected inheritance from grandparents - advice? & Legal-Advice & Financial-Advice\\
    \hline
    \hline
    \end{tabular}
        \caption{Examples of error made by different approaches with the best performing model. Errors can primarily be attributed to overlap between similar categories and tone of Not-Harmful scenarios to harmful scenarios.}

    \label{tab:error-example}
\end{table*}

% We apply knowledge distillation from GPT-4 with the use of a template to generate variations.
% \setlength{\tabcolsep}{10pt}

   \begin{table*}[h]
    \centering
    \small
    \begin{tabular}{l|c|c|c|c}
    \hline
	& \textbf{Random + LLM} & \textbf{Topn-AL + LLM} & \textbf{Coreset-AL + LLM} & \textbf{Cluster-AL + LLM}\\
 \hline
\textbf{Stand Deviation of Error $\downarrow$} & 33.75 &	33.22 &	33.39 &	\textbf{29.71}\\

    \hline
    \end{tabular}
        \caption{Standard deviation of  errors across all classes on the full test set, normalized by the class frequency. Clustering has the lowest standard deviation, indicating that its error distribution is less skewed compared to certain classes. This suggests the model is fairer across different groups in the data.}

    \label{tab:error-dev}
\end{table*}
\paragraph{Baseline classification}
As expected, most models perform poorly in this setting, with roberta-base achieving the highest F1 score of 61.6, followed by F1 score of 57.1 by distillbert-base-uncased (Table \ref{tab:stage1-classifier}). Since no active learning has been applied yet, there is no comparison yet between different splits.

\paragraph{Active learning without LLM generation} Among different active learning approaches, clustering-based active learning outperforms others in Table \ref{tab:stage2-classifier}. However, this improvement is not uniform. We can see an improvement anywhere between 0.1\% to 6.5\% compared to random sampling. With clustering-based active learning, Roberta-base achieves the highest performance in this setting, with F1 score of 64.3 \textemdash an improvement of 2.7 compared to baseline classification. Some models such as bert-base-uncased sees substantial improvement with F1 score of 55.8 compared to F1 score of 36.7 in baseline classification. This indicates most models are becoming stable at this stage.

\paragraph{Active learning with LLM generation} From Table \ref{tab:stage3-classifier}, we observe that incorporating LLM generation substantially improves performance. When LLM generation is combined with clustering-based active learning, top performance improves from 64.3 to 71.5 F1 score with roberta-base, outperforming random sampling (66.0), TopN (68.2) and Coreset (66.3) counterparts. This pattern can be observed across other models as well. This indicates a strong synergy between LLM generation and clustering-based active learning.

\paragraph{Transferability of Acquired Data}
Our results also show that data acquired by active learning paradigms are transferable to other models. While a bert-base-cased model was used as the learner model to provide feedback for LLM generation, we see improvement for most transformer models across Tables \ref{tab:stage2-classifier} and \ref{tab:stage3-classifier} when fine-tuned with the same generated data. In particular, the highest F1-score of \textbf{71.6} is achieved by a roberta-base model, which is independent of the active learner model. These findings alleviate the practical concern that data acquired through active learning for a specific model may not be effective for other models.

% We also see improvement for SVM models, which are completely different from the transformer models. However, the improvements are smaller. This can be attributed to the fact that SVMs in general however perform poorly on our data splits due to their inability to predict successfully from small amount of data.

\subsection{Error Analysis}
We perform error analysis with the best model from earlier, robert-base with different LLM generation approaches, analyzing 100 errors from each of the four approaches. Examples of errors are provided in Table \ref{tab:error-example}. Manual examination of errors reveal following observations:
    
    \begin{figure}[]
    \centering
    \includegraphics[width=0.99\linewidth]{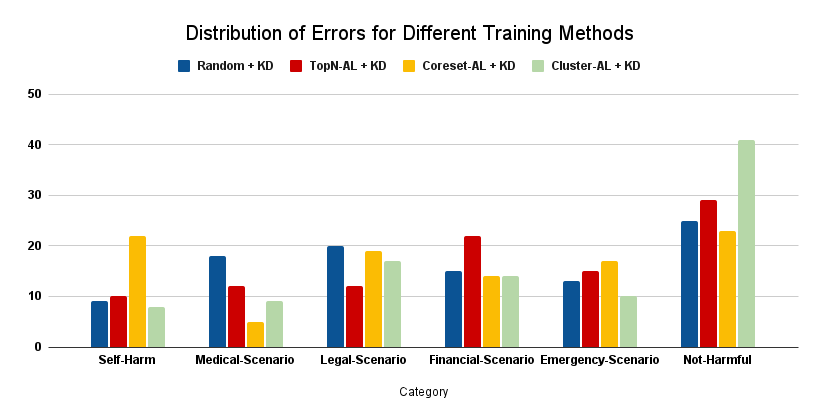}
    \caption{Error distribution across 100 samples, showing more errors in the frequent "Not-Harmful" class and fewer in the under-represented "Emergency Situation" class for our approach. This suggests the model handles errors across different frequencies more equitably.}
    \label{fig:error-dist}
    \end{figure}
\begin{enumerate}
    \item Financial and Legal scenarios can be hard to distinguish due to overlapping concepts.
    \item Words or phrases related to medical advice can be predicted as Medical-Advice even when they are used in benign situations.
    \item Implicit statements of self-harm such as "I've lived everything I want to live" may be hard to categorize as self-harm.
    \item Benign instances that have similar tone to self-harm, may be mis-categorized as self-harm.
\end{enumerate}

Figure \ref{fig:error-dist} shows distribution of these errors. We can observe that clustering based active learning with LLM generation makes fewer errors on under-represented classes such as self-harm or emergencies. When normalized by the number of samples from each class in the full dataset (Table \ref{tab:error-dev}), we observe that clustering-based active learning  has lowest standard deviation of errors across classes, suggesting that our method is more uniform in its errors despite drawing samples from the same unlabeled pool of data. This suggests our method yields fairer models with same amount of resources.

\section{Conclusion and Future Work}
In conclusion, our study proposes a novel framework that integrates active learning and clustering for guiding LLM generation in safety scenarios.  Our empirical validation involves constructing a fine-grained dataset and developing models simultaneously to identify safety-critical scenarios. Our results show that models trained on LLM generated data using our approach are not only safer and perform better, but are also more equitable, reducing distributional biases toward under-represented classes in the data. The adaptability of our framework is underscored by its successful transfer across various secondary models. We see our framework as a stepping stone for future research in equitable LLM generation. We hope our work can encourage the incorporation of clustering-based active learning for generative scenarios such as paraphrasing \cite{atwell-etal-2022-appdia}, responding in sensitive scenarios \cite{hassan-alikhani:2023:ijcnlp}, or within dialogue systems \cite{sicilia2023isabel}.

% We hope that our framework will serve as a catalyst for future research and development in the field. 

\section*{Limitations}
In our work, we outline a framework for guiding LLM generated data with active learning. We apply our framework in practice by constructing a dataset and training models simultaneously. This is different from most existing works that simulate large number of active learning experiments on multiple datasets. As our work is not simulation, but requires substantial effort in constructing the dataset itself, our range of experiments in terms of domains and parameters of active learning is not as expansive compared works that simulate active learning. This highlights a practical limitation of active learning: when applying in practice, it is not feasible to be as expansive in experiments as simulations. Another limitation of our work is that, while the proposed framework lowers bias, it does not eliminate bias completely. Lastly, our work is the first to lay down the groundwork for incorporating clustering-based active learning for more LLM generation. Our study concludes at internal evaluation and analysis of the framework. Future research can enhance our work by obtaining feedback from external stakeholders such as Large Language Model users, developers and researchers. 

% Thus, models trained using our framework should also go through rigorous evaluation and analysis.

\section*{Ethical Considerations}

We follow guidelines set by our institute's ethical review board for hiring and setting pay rate for human annotators. We also follow Reddit's policies \footnote{https://www.redditinc.com/policies/developer-terms} for collecting our unlabeled pool of data. We also follow OpenAI's usage policies \footnote{https://openai.com/policies/usage-policies} for using GPT 3.5.

Our proposed approach allows for more efficient data generation. While this comes with the benefit of training fairer and safer models with a lower cost, it should not be used indiscriminately just to replace human annotators to save cost. Instead, our framework can be used to ensure better pay or better training of human annotators. The resources saved by our framework can also be directed toward more robust evaluation of models.

% Bibliography entries for the entire Anthology, followed by custom entries
%\bibliography{anthology,custom}
% Custom bibliography entries only
\bibliography{custom}

\end{document}